\documentclass[sigconf,screen]{acmart}

\setcopyright{none}
\renewcommand\footnotetextcopyrightpermission[1]{}
\usepackage{microtype}
\usepackage{graphicx}
\usepackage{subcaption}
\usepackage{booktabs} 
\usepackage{multirow} 
\usepackage{adjustbox}
\usepackage{paralist}
\usepackage{hyperref}

\usepackage{amsmath}
\usepackage{amssymb}
\usepackage{mathtools}
\usepackage{amsthm}

\usepackage[capitalize,noabbrev]{cleveref}

\theoremstyle{plain}

\theoremstyle{definition}

\theoremstyle{remark}

\title[Simplicity Prevails for Generalizable AIGI Detection]{Simplicity Prevails: The Emergence of Generalizable\\AIGI Detection in Visual Foundation Models}

\author{Yue Zhou$^{1}$, Xinan He$^{2,1}$, Kaiqing Lin$^{1}$, Bing Fan$^{3}$, Feng Ding$^{2}$, Bin Li$^{1 \ast}$\\
$^{1}$Shenzhen University {\tt \small 2450042008@email.szu.edu.cn;} \\
$^{2}$Nanchang University {\tt \small shahur@email.ncu.edu.cn};\\
$^{3}$University of North Texas} 

\renewcommand{\shortauthors}{Anonymous Author(s)}

\acmConference[ACM MM '26]{Proceedings of the 34th ACM International Conference on Multimedia}{October 2026}{Amsterdam, The Netherlands}
\acmBooktitle{Proceedings of the 34th ACM International Conference on Multimedia (ACM MM '26), October 2026, Amsterdam, The Netherlands}
\acmYear{2026}
\copyrightyear{2026}
\acmDOI{TBD}
\acmISBN{TBD}

\begin{document}

\begin{abstract}
Specialized detectors for AI-generated images (AIGI) often achieve near-perfect accuracy on curated benchmarks, yet their performance degrades substantially in realistic, in-the-wild scenarios. In this work, we show that frozen features from modern Vision Foundation Models (VFMs), combined with a lightweight classifier, form a remarkably strong baseline for generalizable AIGI detection. Using representative modern encoders, including Perception Encoder, MetaCLIP 2, and DINOv3, we conduct a comprehensive evaluation across standard benchmarks, recent unseen generators, and challenging in-the-wild distributions. Across these settings, this simple baseline consistently matches or outperforms recent specialized detectors, with particularly large gains in realistic scenarios.

We further investigate why this simple setup is so effective. Our analyses provide converging evidence that the strong forensic separability of modern VFMs is closely related to their exposure to synthetic web content during pre-training. In Vision-Language Models, this manifests as semantic alignment with forgery-related concepts, while in Self-Supervised Learning models it appears as implicit discrimination of generative distributions. Although a fully controlled pre-training study is beyond the scope of this work, multiple complementary analyses support this interpretation.
We also identify important limitations. While modern VFMs are highly effective for global AIGI detection, they remain vulnerable to severe transmission degradation and perform poorly on pure VAE reconstruction and localized editing. Overall, our results suggest that progress in generalizable AIGI detection may depend more on preserving and leveraging strong pretrained representations than on increasingly complex task-specific forensic designs.
\end{abstract}



\maketitle

\section{Introduction}
\label{sec:intro}

\begin{quote}
\textit{``One thing that should be learned from the bitter lesson is the great power of general purpose methods, of methods that continue to scale with increased computation even as the available computation becomes very great.''} \\
\hspace*{\fill} --- Rich Sutton, \textit{The Bitter Lesson} \cite{sutton2019bitter}
\end{quote}

The rapid evolution of generative models, such as Midjourney, Stable Diffusion \cite{rombach2021high} and Nano Banana\cite{team2023gemini}, has ushered in a new era of content creation, synthesizing photorealistic images that challenge the boundaries of visual authenticity. While empowering creativity, this technological leap simultaneously introduces profound threats to information integrity, fueling the proliferation of misinformation. In response, the forensics community has largely favored a specialized approach: crafting detectors with increasingly complex module tailored to specific artifacts, such as frequency anomalies or noise residuals \cite{wang2020cnn, ju2022fusing}. While these specialized detectors achieve near-perfect accuracy on curated benchmarks, they often suffer from a dramatic performance collapse in realistic scenarios. Recent studies, such as the Chameleon benchmark \cite{yan2024sanity}, reveal that detectors excelling in controlled environments frequently degrade to 60\%--70\% accuracy when deployed `in-the-wild'. This fragility suggests that relying on hand-crafted inductive biases may be a dead end in the face of rapidly evolving generative distributions.

\begin{figure}[t]
    \centering
    \includegraphics[width=\linewidth]{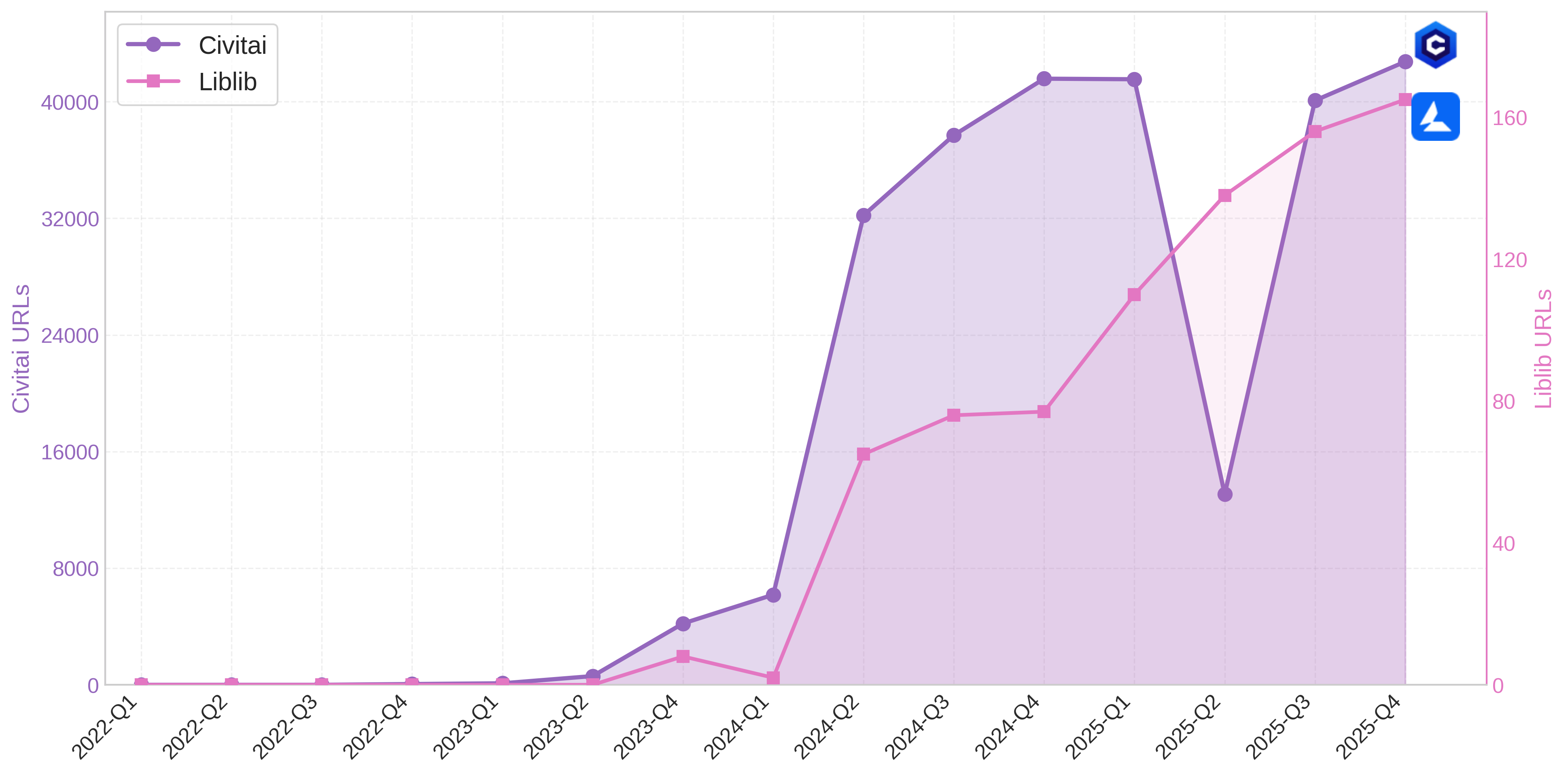}
    \caption{\textbf{The Surge of Generative Data in Web Corpora.} We track the number of indexed URLs from major open-source AI generation platforms (Civitai and Liblib) within Common Crawl snapshots from 2022 to 2025.}
    \label{fig:civitai_trend}
\vspace{-15pt}
\end{figure}

Echoing Sutton's ``Bitter Lesson'', we revisit this trend from a different perspective: rather than asking how to design increasingly specialized forensic modules, we ask how far one can go by directly leveraging the frozen representations of \textbf{modern Vision Foundation Models (VFMs)}. We show that a simple linear classifier, trained on top of frozen features from recent encoders such as Perception Encoder (PE) \cite{bolya2025perception}, MetaCLIP 2 \cite{chuang2025metaclip}, and DINOv3 \cite{simeoni2025dinov3}, provides a remarkably strong baseline for generalizable AIGI detection. We define modern VFMs as the latest generation of encoders trained on large-scale and evolving web corpora. Our evaluation spans three distinct protocols: standard benchmarks (e.g., GenImage~\cite{zhu2023genimage}), datasets from the latest unseen generators (e.g., AIGIHolmes~\cite{zhou2025aigi}, AIGI-Now~\cite{chen2025task}), and challenging in-the-wild distributions (e.g., Chameleon~\cite{yan2024sanity}, WildRF~\cite{cavia2024real}). Across all settings, this simple baseline consistently matches or outperforms recent specialized detectors, with the largest gains appearing in the most challenging in-the-wild scenarios.

We further investigate why such a simple setup is so effective. Rather than attributing the gains to forensic-specific architectural innovation, we argue that they are closely related to an \textbf{emergent property} of large-scale pre-training on evolving web data. As visualized in Figure~\ref{fig:civitai_trend}, our analysis of the Common Crawl index reveals a rapid increase in generative content from major communities such as Civitai and Liblib starting in 2023. This trend suggests that modern VFMs are increasingly likely to encounter synthetic content during pre-training, and may therefore internalize useful cues for distinguishing generated images from real ones. We characterize this capability through two distinct manifestations of data exposure. For Vision-Language Models, the co-occurrence of synthetic images and textual descriptions can lead to \textbf{explicit concept injection}, where synthetic visuals become aligned with high-level forgery-related concepts. For Self-Supervised Learning (SSL) models such as DINOv3, the capability appears instead as \textbf{implicit distribution fitting}, where the model captures low-level regularities associated with the generative manifold through pre-training data exposure.

At the same time, our analysis also clarifies the boundaries of this paradigm. We find that modern VFMs remain \textbf{blind to pure reconstruction artifacts} and struggle with \textbf{localized editing}, while also suffering from noticeable degradation under aggressive real-world transmission and screen recapture. These findings suggest that large-scale pre-training has substantially improved the \textit{generalizability} of global detection for fully synthetic content, but that the \textit{localization} of fine-grained manipulation remains an open challenge. Complementing these observations, our experiments on \textbf{backbone replacement} and \textbf{lightweight LoRA fine-tuning} suggest that the dominant factor behind performance is still the strength of the pretrained representation itself, while adaptation mainly serves to better exploit this foundation. Therefore, rather than viewing increasingly specialized detector design as the default path forward, we argue that future progress in AI forensics may depend more on how to preserve, harness, and refine the evolving representations of foundation models for fine-grained forensic reasoning.

In summary, our main contributions are:
\begin{itemize}
\item We show that frozen features from modern Vision Foundation Models, combined with a lightweight classifier, constitute a remarkably strong baseline for generalizable AIGI detection. Across standard benchmarks, recent unseen generators, and challenging in-the-wild distributions, this simple setup consistently matches or outperforms recent specialized detectors, with especially large gains in realistic scenarios.
\item We provide converging evidence that this capability is closely related to pre-training data exposure rather than forensic-specific architectural design. In particular, we identify two complementary manifestations: semantic alignment with forgery-related concepts in Vision-Language Models, and implicit discrimination of generative distributions in Self-Supervised Learning models.
\item We validate the \textbf{``Bitter Lesson''} in AI forensics and delineate the \textbf{boundaries of this paradigm}. Through rigorous ablation, we demonstrate that attaching complex forensic heads or applying fine-tuning (e.g., LoRA) actively degrades the generic representations of VFMs. While these frozen generic features solve global detection, they remain blind to pure VAE reconstruction and localized editing, urging future research to harness rather than over-engineer foundation models.
\end{itemize}

\begin{table*}[ht]
\centering
\caption{\textbf{Performance on GenImage Benchmark.} All detectors are trained on Stable Diffusion v1.4 and evaluated on unseen generators. Accuracy is averaged over real and fake classes. Best results in \textbf{bold}.}
\label{tab:genimage}
\small
\resizebox{0.9\textwidth}{!}{
\begin{tabular}{l|cccccccc|c}
\toprule
\textbf{Method} & \textbf{ADM} & \textbf{BigGAN} & \textbf{Midjourney} & \textbf{VQDM} & \textbf{GLIDE} & \textbf{SD-v1.4} & \textbf{SD-v1.5} & \textbf{Wukong} & \textbf{Avg.} \\
\midrule
\multicolumn{10}{l}{\textit{Modern VFM Baselines (Ours)}} \\
MetaCLIP-Linear & 0.549 & 0.524 & 0.839 & 0.653 & 0.653 & 0.984 & 0.981 & 0.942 & 0.766 \\
MetaCLIP2-Linear & 0.690 & 0.816 & 0.959 & 0.819 & 0.887 & 0.993 & 0.991 & 0.980 & 0.892 \\
SigLIP-Linear & 0.740 & 0.658 & 0.812 & 0.857 & 0.916 & 0.955 & 0.954 & 0.918 & 0.851 \\
SigLIP2-Linear & 0.870 & 0.924 & 0.879 & 0.954 & 0.951 & 0.995 & 0.995 & 0.994 & 0.945 \\
PE-CLIP-Linear & 0.712 & 0.963 & 0.901 & 0.959 & 0.972 & \textbf{0.999} & \textbf{0.999} & \textbf{0.999} & 0.938 \\
DINOv2-Linear & 0.651 & 0.889 & 0.811 & 0.757 & 0.852 & 0.967 & 0.961 & 0.929 & 0.852\\
DINOv3-Linear & 0.849 & \textbf{0.991} & 0.934 & \textbf{0.992} & 0.963 & 0.998 & 0.996 & 0.994 & \textbf{0.964} \\
\midrule
\multicolumn{10}{l}{\textit{Competitor Methods}} \\
CNNSpot \cite{wang2020cnn} & 0.507 & 0.530 & 0.610 & 0.501 & 0.522 & 0.998 & 0.996 & 0.985 & 0.706 \\
FreqNet \cite{tan2024frequency} & 0.608 & 0.939 & 0.849 & 0.626 & 0.858 & 0.950 & 0.949 & 0.937 & 0.840 \\
Gram-Net \cite{liu2020global} & 0.576 & 0.591 & 0.665 & 0.528 & 0.747 & 0.989 & 0.988 & 0.960 & 0.756 \\
NPR \cite{tan2024rethinking} & 0.619 & 0.849 & 0.850 & 0.567 & 0.844 & 0.996 & 0.994 & 0.984 & 0.838 \\
UnivFD \cite{ojha2023towards} & 0.571 & 0.839 & 0.795 & 0.644 & 0.842 & 0.959 & 0.958 & 0.926 & 0.817 \\
SAFE \cite{li2025improving} & 0.600 & 0.880 & 0.829 & 0.708 & 0.836 & 0.964 & 0.967 & 0.903 & 0.836 \\
LaDeDa \cite{cavia2024real} & 0.512 & 0.583 & 0.595 & 0.513 & 0.585 & 0.876 & 0.871 & 0.820 & 0.670 \\
Effort \cite{yan2024orthogonal} & 0.704 & 0.749 & 0.756 & 0.824 & 0.807 & 0.874 & 0.873 & 0.867 & 0.807 \\
DDA \cite{chen2025dual} & \textbf{0.880} & 0.741 & \textbf{0.960} & 0.719 & 0.862 & 0.986 & 0.985 & 0.986 & 0.890 \\
OMAT \cite{zhou2025breaking} & 0.837 & 0.973 & 0.903 & 0.954 & \textbf{0.974} & 0.974 & 0.973 & 0.975 & 0.946 \\
AIDE \cite{yan2024sanity} & 0.602 & 0.648 & 0.813 & 0.694 & 0.514 & 0.959 & 0.957 & 0.954 & 0.768 \\
\bottomrule
\end{tabular}
}
\end{table*}

\section{Related Works}
\label{sec:evolution}

The development of AI-generated image (AIGI) detection has undergone a significant paradigm shift, evolving from hand-crafted artifact analysis to the adaptation of large-scale foundation models.

\textbf{Early Artifact-Based Detection.}
Initial forensic methods focused on identifying the low-level imperfections inherent to early generative architectures. Researchers found that upsampling operations in GANs and CNNs often leave distinct footprints, such as checkerboard artifacts in the pixel space \cite{odena2016deconvolution} or spectral anomalies in the frequency domain \cite{frank2020leveraging}. Others exploited inconsistencies in color statistics \cite{mccloskey2019detecting} or noise residuals \cite{cozzolino2019noiseprint} to distinguish synthetic content. While effective against specific generators, these hand-crafted features proved brittle against the rapid evolution of generative models, particularly with the advent of Diffusion Models which exhibit fundamentally different artifact patterns.

\textbf{Data-Driven Specialized Detectors.}
With the dominance of Diffusion Models, the focus shifted from detecting CNN-specific upsampling artifacts (e.g., checkerboard patterns in GANs \cite{wang2020cnn}) to identifying the unique traces of the diffusion process.
Researchers proposed reconstructing input images to isolate generative errors: DIRE \cite{wang2023dire} leverages the reconstruction residual from a pre-trained diffusion model as a forensic signal, while DRCT \cite{chen2024drct} refines this by analyzing the discrepancies between real-real and fake-fake reconstruction pairs.
Others focused on improving generalization: SAFE \cite{li2025improving} introduces artifact-preserving augmentations to decouple semantic content from forensic traces. notably, DDA \cite{chen2025dual} targets the shared VAE decoder inherent to Latent Diffusion Models, explicitly aligning the detector with VAE reconstruction patterns to achieve broader generalization across different LDM-based generators.

\textbf{The Foundation Model Era.}
The introduction of UnivFD~~\cite{ojha2023towards} marked a pivotal turning point. Ojha et al. revealed that training a linear layer on top of the frozen feature space of a pre-trained Vision-Language Model (specifically CLIP \cite{radford2021learning}) yields significantly better generalization than training CNNs from scratch. This discovery spurred a wave of research leveraging Vision Foundation Models (VFMs) as backbones. Subsequent works, such as Effort\cite{yan2024orthogonal}, AIDE \cite{yan2024sanity}, OMAT\cite{zhou2025breaking} and DDA\cite{chen2025dual}, have explored various strategies to adapt CLIP or DINOv2 \cite{oquab2023dinov2} for forensics, including prompt tuning, adapter modules, fusing frequency-domain information and training hard samples. These methods currently represent the state-of-the-art, yet as our experiments show, they still struggle to maintain robustness in unconstrained, in-the-wild scenarios compared to the raw capabilities of the latest VFMs. (See Appendix for a comprehensive survey of the legacy backbones still prevalent in these recent methods.)

\section{Simplicity Prevails: Benchmarking Modern VFMs}
To empirically validate our hypothesis that the generalization capability of AIGI detection stems from the scale of pre-training data rather than complex architectural designs, we conduct a comprehensive comparative analysis. We pit simple linear classifiers trained on modern Vision Foundation Models (VFMs) against a wide array of state-of-the-art specialized detectors. Our evaluation protocol is designed to be rigorous and progressively challenging, spanning three distinct scenarios: standard academic benchmarks, realistic in-the-wild distributions, and unseen next-generation generative models.

\subsection{Experimental Setup}
\label{sec:setup}

\textbf{Evaluation Benchmarks.} 
To rigorously assess generalization, we organize our evaluation into three progressively challenging categories. 
(1) \textbf{Standard Benchmarks:} We use \textbf{GenImage} \cite{zhu2023genimage}, a widely adopted benchmark comprising images from 8 generators (e.g., Stable Diffusion, Midjourney). Following standard protocols \cite{ojha2023towards}, we use the Stable Diffusion v1.4 subset for training and the remaining subsets for testing.
(2) \textbf{In-the-Wild Datasets:} We evaluate on \textbf{Chameleon} \cite{yan2024sanity}, \textbf{WildRF} \cite{cavia2024real}, \textbf{SocialRF} and \textbf{CommunityAI}~\cite{li2025artificial}. These datasets are collected from social media and internet forums, featuring diverse, unconstrained post-processing and unknown generative sources, representing a realistic detection scenario.
(3) \textbf{Unseen Generators:} We employ \textbf{AIGIHolmes}~\cite{zhou2025aigi} and \textbf{AIGI-Now}~\cite{chen2025task}, recent benchmarks containing images from state-of-the-art generators released after 2024, including closed-source generators like Nano Banana, GPT4o and FLUX-Pro. These serve as a strict test for generalization to unseen distributions.

\textbf{Specialized Detectors.} 
We compare against a comprehensive suite of state-of-the-art forensic methods, spanning three categories: 
(1) \textbf{Specialized Detectors}: Artifact-based methods like CNNSpot \cite{wang2020cnn}, FreqNet \cite{tan2024frequency}, and NPR \cite{tan2024rethinking}, as well as recent VFM-based adapters such as UnivFD \cite{ojha2023towards}, OMAT \cite{zhou2025breaking}, Effort \cite{yan2024orthogonal} and Dual-Data-Alignment (DDA)\cite{chen2025dual}.
(2) \textbf{Former VFM Baselines}: To explicitly evaluate the impact of pre-training data evolution, we include earlier generations of foundation models, such as the original OpenAI CLIP \cite{radford2021learning}, SigLIP~\cite{zhai2023sigmoid}, Meta CLIP~\cite{xu2023demystifying} and DINOv2~\cite{oquab2023dinov2}. For DDA, we utilize the official pre-trained weights provided by the authors, as its core contribution involves a specialized training pipeline with VAE-reconstructed data alignment. All other methods are trained on GenImage SDv1.4 training set for fair comparison.

\textbf{Modern VFM Baselines.} 
To test our ``Simplicity Prevails'' hypothesis, we select a representative set of modern Vision Foundation Models as frozen feature extractors. These include \textbf{Vision-Language Models} (SigLIP2 \cite{tschannen2025siglip}, MetaCLIP 2 \cite{chuang2025metaclip}, Perception Encoder \cite{bolya2025perception}) and \textbf{Self-Supervised Models} (DINOv3 \cite{simeoni2025dinov3}). We attach a simple linear layer to the pooled output features of these backbones. Detailed specifications of model architectures and pre-training datasets are provided in Appendix.

\begin{table*}[ht]
\centering
\caption{\textbf{Performance on In-the-Wild Benchmarks.} Evaluation on Chameleon, WildRF, SocialRF, and CommunityAI datasets. Accuracy is averaged over real and fake classes. Best results in \textbf{bold}.}
\small
\resizebox{0.9\textwidth}{!}{
\begin{tabular}{l|ccc|ccc|ccc|ccc|c}
\toprule
\multirow{2}{*}{\textbf{Method}} & \multicolumn{3}{c|}{\textbf{Chameleon}} & \multicolumn{3}{c|}{\textbf{WildRF}} & \multicolumn{3}{c|}{\textbf{SocialRF}} & \multicolumn{3}{c|}{\textbf{CommunityAI}} & \multirow{2}{*}{\textbf{Avg.}} \\
 & Real & Fake & Avg. & Real & Fake & Avg. & Real & Fake & Avg. & Real & Fake & Avg. & \\
\midrule
\multicolumn{14}{l}{\textit{Modern VFM Baselines (Ours)}} \\
MetaCLIP-Linear & 0.373 & 0.914 & 0.644 & 0.461 & 0.923 & 0.692 & 0.409 & 0.866 & 0.638 & 0.353 & 0.933 & 0.643 & 0.654 \\
MetaCLIP2-Linear & 0.948 & 0.913 & 0.930 & 0.478 & 0.979 & 0.728 & 0.659 & 0.940 & 0.800 & 0.926 & 0.954 & 0.940 & 0.842 \\
SigLIP-Linear & 0.480 & 0.732 & 0.606 & 0.383 & 0.897 & 0.640 & 0.549 & 0.613 & 0.581 & 0.370 & 0.857 & 0.614 & 0.610 \\
SigLIP2-Linear & 0.884 & 0.833 & 0.859 & 0.597 & 0.984 & 0.790 & 0.744 & 0.866 & 0.805 & 0.826 & 0.905 & 0.866 & 0.822 \\
PE-CLIP-Linear & 0.970 & \textbf{0.948} & \textbf{0.959} & 0.679 & \textbf{0.994} & 0.836 & 0.751 & \textbf{0.970} & 0.861 & 0.966 & \textbf{0.975} & \textbf{0.971} & 0.899 \\
DINOv2-Linear & 0.628 & 0.580 & 0.608 & 0.643 & 0.772 & 0.705 & 0.603 & 0.695 & 0.649 & 0.606 & 0.562 & 0.583 & 0.636\\
DINOv3-Linear & 0.933 & 0.895 & 0.914 & 0.948 & 0.975 & \textbf{0.961} & \textbf{0.937} & 0.948 & \textbf{0.943} & 0.949 & 0.946 & 0.948 & \textbf{0.940} \\
\midrule
\multicolumn{14}{l}{\textit{Competitor Methods}} \\
CNNSpot & 0.979 & 0.128 & 0.554 & 0.959 & 0.290 & 0.625 & 0.588 & 0.541 & 0.565 & 0.969 & 0.112 & 0.541 & 0.571 \\
FreqNet & 0.985 & 0.090 & 0.538 & 0.731 & 0.559 & 0.645 & 0.544 & 0.553 & 0.549 & 0.977 & 0.128 & 0.553 & 0.571 \\
Gram-Net & 0.992 & 0.044 & 0.518 & 0.947 & 0.205 & 0.576 & 0.531 & 0.523 & 0.527 & 0.985 & 0.061 & 0.523 & 0.536 \\
NPR & \textbf{0.999} & 0.046 & 0.523 & \textbf{0.980} & 0.243 & 0.612 & 0.593 & 0.537 & 0.565 & \textbf{0.998} & 0.076 & 0.537 & 0.560 \\
UnivFD & 0.763 & 0.441 & 0.602 & 0.693 & 0.637 & 0.665 & 0.563 & 0.637 & 0.600 & 0.778 & 0.495 & 0.636 & 0.617 \\
SAFE & 0.993 & 0.046 & 0.520 & 0.918 & 0.309 & 0.613 & 0.564 & 0.532 & 0.548 & 0.984 & 0.080 & 0.532 & 0.556 \\
LaDeDa & 0.994 & 0.015 & 0.504 & 0.988 & 0.119 & 0.554 & 0.542 & 0.506 & 0.524 & 0.986 & 0.026 & 0.506 & 0.523 \\
Effort & 0.394 & 0.782 & 0.588 & 0.179 & 0.955 & 0.567 & 0.513 & 0.533 & 0.523 & 0.225 & 0.840 & 0.533 & 0.553 \\
DDA & 0.940 & 0.708 & 0.824 & 0.899 & 0.908 & 0.904 & 0.818 & 0.846 & 0.832 & 0.968 & 0.725 & 0.847 & 0.850 \\
OMAT & 0.899 & 0.359 & 0.629 & 0.633 & 0.715 & 0.674 & 0.581 & 0.633 & 0.607 & 0.853 & 0.414 & 0.634 & 0.636 \\
AIDE & 0.944 & 0.203 & 0.574 & 0.973 & 0.195 & 0.584 & 0.578 & 0.541 & 0.560 & 0.990 & 0.093 & 0.542 & 0.565 \\
\bottomrule
\end{tabular}
}
\label{tab:inthewild}
\end{table*}

\textbf{Implementation Details.} 
All our VFM baselines are trained solely on the \textbf{GenImage (SD v1.4)} training set. We keep the backbone completely frozen and only update the linear head. We use the AdamW optimizer with a learning rate of $1e^{-3}$ and a batch size of 128 for 2 epoch. Images are resized and center-cropped to the native resolution of each model \textbf{without any additional data augmentation}. Notably, PE uses a ViT-L/14 backbone at 336px resolution, which is comparable in scale to several strong VFM-based baselines and recent detectors built on CLIP-L style encoders.

\subsection{Performance on Standard Benchmarks}
We first establish the efficacy of our simple baselines on the standard GenImage benchmark in Table~\ref{tab:genimage}. Despite the simplicity of the linear probe, modern VFMs achieve state-of-the-art performance. DINOv3-Linear reaches the highest average accuracy of 96.5\%, surpassing the best specialized detector OMAT (94.6\%) and significantly outperforming legacy baselines. Notably, we observe a substantial performance leap across VFM generations: DINOv3 improves upon DINOv2 by over 11\%, and MetaCLIP-2 boosts accuracy by 12.6\% compared to its predecessor MetaCLIP. This trend highlights that forensic discriminability is not static but scales with the quality and data volume of the foundation model. Furthermore, while specialized detectors often overfit to the training source, modern VFMs demonstrate robust generalization across diverse generative architectures, confirming that their representations are inherently forensic-ready without the need for complex auxiliary modules.

\subsection{The Collapse of SOTA in the Wild}
While standard benchmarks provide a controlled environment, real-world deployment involves diverse, unconstrained data distributions. To evaluate this, we test on four challenging in-the-wild datasets: \textbf{Chameleon}, \textbf{WildRF}, \textbf{Social-RF}, and \textbf{CommunityAI}. The results, summarized in Table~\ref{tab:inthewild}, reveal a stark contrast.
Most specialized detectors suffer a catastrophic performance collapse. Methods like NPR, LaDeDa, and even recent techniques like OMAT degrade to near-random performance, primarily due to a failure to recognize diverse fake samples. The only exception among specialized methods is DDA, which maintains a respectable average accuracy of 85.0\%, due to its alignment with the robust VAE decoder patterns shared across latent diffusion models by training on DINOv2 with VAE reconstruct data.

However, modern VFM baselines decisively outperform all competitors. DINOv3-Linear achieves an average accuracy of 94.0\%, surpassing DDA by nearly 10\% and traditional detectors by over 30\%. Crucially, we observe a massive performance gap between modern and legacy VFMs: DINOv3 outperforms its predecessor DINOv2 by a staggering 30.4\%, and MetaCLIP-2 surpasses MetaCLIP by 18.8\%. This confirms that earlier foundation models lacked the necessary data exposure to handle in-the-wild shifts, whereas modern iterations have internalized these distributions during pre-training.

\begin{table*}[h]
\centering
\caption{\textbf{Generalization on AIGIHolmes.} Evaluation on advanced auto-regressive and diffusion-transformer generative models. Accuracy is averaged over real and fake classes. Best results in \textbf{bold}.}
\small
\resizebox{\textwidth}{!}{
\begin{tabular}{l|c|c|c|c|c|c|c|c|c|c|c}
\toprule
\textbf{Detector} &
\textbf{FLUX} &
\textbf{Infinity} &
\textbf{Janus} &
\textbf{Janus-Pro-1B} &
\textbf{Janus-Pro-7B} &
\textbf{LlamaGen} &
\textbf{PixArt-XL} &
\textbf{SD3.5-L} &
\textbf{Show-o} &
\textbf{VAR} &
\textbf{Avg} \\
\midrule
\multicolumn{12}{l}{\textit{Modern VFM Baselines (Ours)}} \\
MetaCLIP-Linear & 0.951 & 0.978 & 0.736 & 0.950 & 0.950 & 0.970 & 0.970 & 0.867 & 0.972 & 0.614 & 0.896 \\
MetaCLIP2-Linear & 0.987 & 0.990 & 0.839 & 0.959 & 0.928 & 0.989 & 0.986 & 0.956 & 0.985 & 0.802 & 0.942 \\
SigLIP-Linear & 0.888 & 0.918 & 0.889 & 0.928 & 0.926 & 0.913 & 0.896 & 0.826 & 0.868 & 0.834 & 0.889 \\
SigLIP2-Linear & 0.957 & 0.994 & 0.990 & 0.993 & 0.989 & 0.991 & 0.994 & 0.914 & 0.992 & 0.913 & 0.973 \\
PE-CLIP-Linear & 0.968 & \textbf{0.999} & 0.945 & 0.995 & \textbf{0.996} & \textbf{1.000} & \textbf{1.000} & 0.943 & \textbf{0.999} & 0.935 & \textbf{0.978} \\
DINOv3-Linear & 0.933 & 0.998 & \textbf{0.996} & \textbf{0.995} & 0.986 & 0.999 & 0.999 & 0.891 & 0.997 & 0.922 & 0.972 \\
\midrule
\multicolumn{12}{l}{\textit{Competitor Methods}} \\
CNNSpot & 0.626 & 0.589 & 0.508 & 0.531 & 0.522 & 0.604 & 0.668 & 0.568 & 0.608 & 0.504 & 0.573 \\
FreqNet & 0.894 & 0.924 & 0.443 & 0.439 & 0.433 & 0.875 & 0.885 & 0.894 & 0.910 & 0.688 & 0.738 \\
Gram-Net & 0.696 & 0.604 & 0.491 & 0.491 & 0.491 & 0.627 & 0.627 & 0.701 & 0.771 & 0.557 & 0.606 \\
NPR & 0.968 & 0.983 & 0.495 & 0.495 & 0.495 & 0.947 & 0.899 & 0.937 & 0.988 & 0.730 & 0.794 \\
LaDeDa & 0.682 & 0.664 & 0.498 & 0.497 & 0.497 & 0.810 & 0.641 & 0.695 & 0.755 & 0.724 & 0.646 \\
UnivFD & 0.868 & 0.898 & 0.575 & 0.674 & 0.575 & 0.915 & 0.925 & 0.898 & 0.911 & 0.597 & 0.784 \\
SAFE & 0.932 & 0.965 & 0.482 & 0.482 & 0.483 & 0.913 & 0.896 & 0.940 & 0.962 & 0.911 & 0.796 \\
Effort-AIGI & 0.794 & 0.804 & 0.483 & 0.650 & 0.569 & 0.798 & 0.804 & 0.765 & 0.793 & 0.764 & 0.722 \\
DDA & \textbf{0.972} & 0.989 & 0.985 & 0.993 & 0.987 & 0.993 & 0.994 & 0.970 & 0.948 & 0.804 & 0.963 \\
OMAT & 0.947 & 0.955 & 0.651 & 0.756 & 0.641 & 0.967 & 0.969 & 0.957 & 0.969 & \textbf{0.959} & 0.877 \\
AIDE & 0.944 & 0.987 & 0.912 & 0.989 & 0.978 & 0.994 & 0.986 & \textbf{0.994} & 0.980 & 0.936 & 0.970 \\
\bottomrule
\end{tabular}
}
\label{tab:aigi-holmes}
\end{table*}

\begin{table*}[h]
\centering
\caption{\textbf{Generalization on AIGI-Now.} Evaluation on 9 open-source and closed-source generative models. Accuracy is averaged over real and fake classes. Best results in \textbf{bold}.}
\label{tab:aigi-now}
\small
\resizebox{\textwidth}{!}{
\begin{tabular}{l|cc|cc|cc|cc|cc|cc|cc|cc|cc|c}
\toprule
\multirow{2}{*}{\textbf{Detector}} & 
\multicolumn{2}{c|}{\textbf{FLUX-dev}} & 
\multicolumn{2}{c|}{\textbf{FLUX-kera}} & 
\multicolumn{2}{c|}{\textbf{FLUX-kontext}} & 
\multicolumn{2}{c|}{\textbf{FLUX-pro}} & 
\multicolumn{2}{c|}{\textbf{gpt4o}} & 
\multicolumn{2}{c|}{\textbf{jimeng}} & 
\multicolumn{2}{c|}{\textbf{keling}} & 
\multicolumn{2}{c|}{\textbf{minimax}} & 
\multicolumn{2}{c|}{\textbf{Nano}} & 
\multirow{2}{*}{\textbf{Avg.}} \\
& pix & sem & pix & sem & pix & sem & pix & sem & pix & sem & pix & sem & pix & sem & pix & sem & pix & sem & \\
\midrule
\multicolumn{20}{l}{\textit{Modern VFM Baselines (Ours)}} \\
MetaCLIP-Linear & 0.948 & 0.965 & 0.956 & \textbf{0.913} & 0.708 & 0.808 & 0.894 & 0.943 & 0.898 & 0.882 & 0.846 & 0.829 & 0.963 & \textbf{0.939} & 0.889 & 0.877 & 0.917 & 0.884 & 0.892 \\
MetaCLIP2-Linear & \textbf{0.979} & 0.941 & \textbf{0.963} & 0.896 & 0.799 & \textbf{0.811} & \textbf{0.976} & 0.892 & 0.943 & 0.888 & \textbf{0.965} & 0.825 & 0.970 & 0.902 & \textbf{0.942} & 0.850 & 0.965 & 0.819 & \textbf{0.907} \\
SigLIP-Linear & 0.838 & 0.898 & 0.840 & 0.891 & 0.700 & \textbf{0.811} & 0.836 & 0.911 & 0.827 & 0.897 & 0.827 & 0.890 & 0.863 & 0.937 & 0.848 & 0.863 & 0.796 & 0.867 & 0.852 \\
SigLIP2-Linear & 0.947 & 0.882 & 0.883 & 0.697 & 0.776 & 0.678 & 0.888 & 0.885 & 0.936 & 0.790 & 0.831 & 0.845 & 0.941 & 0.867 & 0.850 & 0.688 & 0.895 & 0.882 & 0.843 \\
PE-CLIP-Linear & 0.977 & 0.959 & 0.918 & 0.762 & 0.830 & 0.774 & 0.873 & 0.943 & 0.863 & 0.924 & 0.915 & 0.921 & 0.939 & 0.916 & 0.865 & 0.748 & 0.971 & 0.936 & 0.891 \\
DINOv3-Linear & 0.944 & \textbf{0.962} & 0.846 & 0.811 & 0.730 & 0.756 & 0.813 & \textbf{0.948} & 0.898 & \textbf{0.960} & 0.824 & \textbf{0.940} & 0.884 & 0.913 & 0.727 & 0.784 & 0.898 & \textbf{0.922} & 0.864 \\
\midrule
\multicolumn{20}{l}{\textit{Competitor Methods}} \\
CNNSpot & 0.919 & 0.500 & 0.550 & 0.500 & 0.843 & 0.502 & 0.535 & 0.500 & \textbf{0.990} & 0.501 & 0.523 & 0.500 & \textbf{0.973} & 0.501 & 0.603 & 0.504 & 0.985 & 0.499 & 0.635 \\
FreqNet & 0.875 & 0.468 & 0.697 & 0.443 & 0.769 & 0.479 & 0.492 & 0.501 & 0.923 & 0.510 & 0.459 & 0.530 & 0.922 & 0.543 & 0.824 & 0.511 & 0.907 & 0.487 & 0.622 \\
Gram-Net & 0.933 & 0.528 & 0.667 & 0.522 & 0.864 & 0.555 & 0.608 & 0.569 & 0.763 & 0.508 & 0.508 & 0.503 & 0.955 & 0.554 & 0.717 & 0.521 & 0.905 & 0.528 & 0.651 \\
NPR & 0.944 & 0.500 & 0.508 & 0.500 & 0.785 & 0.502 & 0.502 & 0.502 & 0.966 & 0.500 & 0.497 & 0.501 & 0.957 & 0.500 & 0.548 & 0.500 & 0.930 & 0.500 & 0.619 \\
LaDeDa & 0.586 & 0.498 & 0.497 & 0.497 & 0.560 & 0.502 & 0.496 & 0.495 & 0.745 & 0.499 & 0.495 & 0.497 & 0.661 & 0.501 & 0.509 & 0.502 & 0.766 & 0.505 & 0.546 \\
UnivFD & 0.542 & 0.579 & 0.514 & 0.538 & 0.492 & 0.545 & 0.516 & 0.544 & 0.529 & 0.532 & 0.475 & 0.504 & 0.631 & 0.539 & 0.501 & 0.539 & 0.501 & 0.516 & 0.531 \\
SAFE & 0.903 & 0.490 & 0.532 & 0.492 & 0.831 & 0.494 & 0.521 & 0.486 & 0.977 & 0.488 & 0.509 & 0.486 & 0.960 & 0.487 & 0.590 & 0.491 & 0.961 & 0.484 & 0.621 \\
Effort-AIGI & 0.789 & 0.679 & 0.796 & 0.669 & 0.728 & 0.610 & 0.688 & 0.690 & 0.753 & 0.580 & 0.522 & 0.555 & 0.782 & 0.677 & 0.772 & 0.687 & 0.796 & 0.657 & 0.690 \\
DDA & 0.916 & 0.512 & 0.594 & 0.499 & 0.827 & 0.529 & 0.766 & 0.550 & 0.923 & 0.654 & 0.870 & 0.654 & 0.961 & 0.646 & 0.833 & 0.505 & 0.816 & 0.562 & 0.695 \\
OMAT & 0.911 & 0.475 & 0.649 & 0.469 & 0.847 & 0.507 & 0.591 & 0.515 & 0.744 & 0.452 & 0.491 & 0.465 & 0.936 & 0.526 & 0.699 & 0.467 & 0.891 & 0.468 & 0.615 \\
AIDE & 0.991 & 0.590 & 0.504 & 0.569 & \textbf{0.979} & 0.806 & 0.601 & 0.538 & 0.747 & 0.518 & 0.639 & 0.514 & 0.982 & 0.554 & 0.514 & 0.541 & \textbf{0.989} & 0.518 & 0.672 \\
\bottomrule
\end{tabular}
}
\end{table*}

\subsection{Generalization to State-of-the-Art Generators}
\label{sec:generalization}

A critical question remains: do these models truly learn generalized forensic concepts, or do they merely memorize pre-training patterns? We investigate this via \textbf{AIGIHolmes} and \textbf{AIGI-Now}, which challenge detectors across two fronts:

\begin{compactitem}
 \item \textbf{AIGIHolmes:} Features recent Auto-Regressive (AR) models (e.g., LlamaGen, VAR) and Diffusion Transformers (e.g., FLUX), whose mechanisms differ fundamentally from older UNet-based models (e.g., SD-v1.4).
    \item \textbf{AIGI-Now:} Contains closed-source APIs (e.g., GPT-4o, FLUX-Pro) unseen during VFM pre-training. And disentangles evaluation into two subsets: \textbf{Pixel-artifact (pix)}, which isolates low-level generative traces by strictly aligning image formats; and \textbf{Semantic (sem)}, which applies aggressive degradations to obliterate low-level artifacts, forcing detectors to rely solely on high-level semantic anomalies.
\end{compactitem}

\textbf{Results.} As shown in Tables~\ref{tab:aigi-holmes} and \ref{tab:aigi-now}, modern VFMs demonstrate exceptional transferability. On AIGIHolmes, PE-CLIP and DINOv3 achieve average accuracies of \textbf{97.8\%} and \textbf{97.2\%}, maintaining robust detection even on fundamentally distinct AR models (e.g., VAR). 

On AIGI-Now, MetaCLIP2 leads with \textbf{90.7\%}. Crucially, VFMs excel across both the \textbf{pix} and heavily degraded \textbf{sem} subsets, proving they capture a dual-level (structural and semantic) notion of artificiality. Conversely, specialized detectors like CNNSpot collapse to near-random guessing ($\sim$50\%) on the \textbf{sem} splits. This confirms that modern VFMs learn generalized, robust forensic concepts that extend far beyond specific generators or brittle low-level artifacts.
\begin{table*}[th!]
\centering
\caption{\textbf{Comparison of Text--Image Similarities on In-the-Wild Dataset}}
\label{tab:textimgmatch}
\begin{adjustbox}{width=0.95\linewidth}
  \begin{tabular}{l|cccccccc}
  \toprule
  \multirow{3}{*}{Method} & \multicolumn{4}{c}{Chameleon}& \multicolumn{4}{c}{SocialRF}\\
  \cmidrule(r){2-5} \cmidrule(r){6-9}
   & \multicolumn{2}{c}{Top-1} & \multicolumn{2}{c}{Top-2} & \multicolumn{2}{c}{Top-1} & \multicolumn{2}{c}{Top-2} \\
   \cmidrule(r){2-3} \cmidrule(r){4-5} \cmidrule{6-7} \cmidrule{8-9}
   &Matched Text& Similarity Score&Matched Text& Similarity Score&Matched Text& Similarity Score&Matched Text& Similarity Score\\
  \midrule

    CLIP \textsubscript{(2021.2.26)} &modern design&0.628&portrait&0.345&forged&0.332&urban&0.253\\

  Siglip \textsubscript{(2023.3.27)}&unaltered&0.318&original&0.212&edited&0.232&original&0.222\\
  Siglip2 \textsubscript{(2025.2.21)}&genuine&0.385&urban&0.250&portrait&0.202&vintage&0.191\\
  \textcolor{blue}{Meta CLIP} \textsubscript{\textcolor{blue}{(2023.9.28)}}&AI generated&0.678&deepfake&0.091&AI generated&0.902&original&0.024\\
  \textcolor{blue}{Meta CLIP-2} \textsubscript{\textcolor{blue}{(2025.7.29)}} &AI generated&0.828&deepfake&0.064&AI generated&0.924&deepfake&0.038\\
  \textcolor{blue}{PE} \textsubscript{\textcolor{blue}{(2025.4.17)}}&AI generated&0.861&deepfake&0.021&AI generated&0.943&deepfake&0.031\\

  \bottomrule
  \multirow{3}{*}{Method} & \multicolumn{4}{c}{CommunityAI}& \multicolumn{4}{c}{Midjourney-CC} \\
  \cmidrule(r){2-5} \cmidrule(r){6-9}
   & \multicolumn{2}{c}{Top-1} & \multicolumn{2}{c}{Top-2} & \multicolumn{2}{c}{Top-1} & \multicolumn{2}{c}{Top-2} \\
      \cmidrule(r){2-3} \cmidrule(r){4-5} \cmidrule{6-7} \cmidrule{8-9}

&Matched Text& Similarity Score&Matched Text& Similarity Score&Matched Text& Similarity Score&Matched Text& Similarity Score\\
    \midrule

  CLIP \textsubscript{(2021.2.26)}&portrait&0.346&nature&0.326&urban&0.332&midjourney\_images&0.260 \\
  Siglip \textsubscript{(2023.3.27)}&AIGIBench&0.283&real&0.281&fake&0.278&original&0.246\\
  Siglip2 \textsubscript{(2025.2.21)}&urban&0.209&portrait&0.201&urban&0.212&portrait&0.194\\
  \textcolor{blue}{Meta CLIP} \textsubscript{\textcolor{blue}{(2023.9.28)}}&AI generated&0.726&deepfake&0.086&midjourney\_images&0.604&AI generated&0.284\\
  \textcolor{blue}{Meta CLIP-2} \textsubscript{\textcolor{blue}{(2025.7.29)}}&AI generated&0.858&deepfake&0.041&midjourney\_images&0.621&AI generated&0.308\\
  \textcolor{blue}{PE} \textsubscript{\textcolor{blue}{(2025.4.17)}}&AI generated&0.878&CommunityAI&0.029&midjourney\_images&0.722&AI generated&0.218\\

  \bottomrule
   
  \end{tabular}
  \end{adjustbox}
\end{table*}

\section{Analysis: The Mechanisms of Emergence}
\label{sec:analysis}

The strong performance of simple probes on modern VFMs raises a central question: does this capability primarily come from forensic-specific architectural choices, or from properties already present in large-scale pretrained representations? Motivated by the rapid growth of synthetic content on the web (Figure~\ref{fig:civitai_trend}), we investigate whether pre-training data exposure is an important factor behind this phenomenon. Because fully controlled retraining of proprietary billion-parameter models is computationally infeasible, we do not attempt to make a definitive causal claim. Instead, we use a set of complementary indirect analyses to characterize how this capability may emerge. Taken together, these analyses suggest two main mechanisms: semantic conceptualization in Vision-Language Models and implicit distribution discrimination in Self-Supervised Learning models.

\subsection{Mechanism I: Semantic Conceptualization in VLMs}
\label{subsec:mechanism_vlm}

For Vision-Language Models (VLMs), we hypothesize that their capability stems from the contrastive pre-training objective. During training, massive volumes of synthetic images co-occur with metadata or captions containing explicit indicators of their source (e.g., \emph{midjourney},\emph{AI generated}). Consequently, the model internalizes a powerful semantic shortcut: it learns to align the visual features of synthetic content directly with forgery-related textual concepts.
To validate this, we conduct a text-image alignment analysis without training any classifier. We probe whether the frozen embedding space of VLMs naturally clusters synthetic images closer to forgery-related prompts. We constructed a comprehensive text pool categorized into three conceptual groups to probe the model's internal associations:
\begin{compactitem}
\item \textbf{Forgery-Related Concepts:} Terms explicitly denoting authenticity or fabrication (e.g., \emph{`fake', `real', `AI generated', `authentic', `manipulated', `synthetic'}).
\item \textbf{Content-Related Concepts:} Neutral descriptions of visual content (e.g., \emph{`sunset', `landscape', `portrait', `abstract art', `technology', `nature'}).
\item \textbf{Source-Related Concepts:} Specific names of generative models or platforms (e.g., \emph{`GenImage', `ADM', `BigGAN', `glide', `Midjourney'}).
\end{compactitem} 

We evaluate the cosine similarity on in-the-wild benchmarks and our newly collected \textbf{Midjourney-CC} dataset (3,000 images from reddit.com/r/midjourney, late 2025) to strictly control for data leakage.

Table~\ref{tab:textimgmatch} presents the results of our semantic probing, revealing a striking dichotomy between legacy and modern VLMs.
Legacy models like CLIP (2021) and SigLIP (2023) exhibit ``forensic blindness'', mapping synthetic images to content terms (e.g., portrait). Notably, even the recently released SigLIP 2 (2025) fails to detect forgery concepts (Top-1: genuine/urban), likely because it relies on the older WebLI dataset~\cite{chen2022pali} curated in 2022, prior to the generative explosion. In sharp contrast, modern VLMs trained on recent web crawls (MetaCLIP 2, PE) consistently align fake images with \textbf{``AI generated''}. Crucially, on Midjourney-CC, these models specifically retrieve \textbf{``midjourney\_images''}, providing definitive evidence that their capability stems from exposure to recent, platform-specific metadata which older datasets lack.

\subsection{Mechanism II: Data-Driven Feature Discrimination in SSL}
\label{subsec:mechanism_ssl}

While VLMs rely on semantic tags, Self-Supervised Learning (SSL) models like DINOv3 lack textual supervision, yet they often outperform VLMs in our benchmarks. We hypothesize that this capability is acquired implicitly through \textbf{distribution fitting}: by training on a massive web corpus mixed with generative content, the model learns to encode the distinct \textbf{low-level signatures of the generative manifold} into its feature space as separable clusters, independent of semantic labels.

To validate that this capability stems from \textbf{data exposure} rather than architectural advantages, we conduct a counterfactual experiment. We employ the identical \textbf{DINOv3 ViT-7B} architecture but vary the pre-training data source: \textbf{DINOv3-Web (LVD-1689M):} Pre-trained on a large-scale web corpus containing 1.6 billion diverse internet images, which naturally includes a significant volume of AIGI. \textbf{DINOv3-Sat (Sat-493M):} Pre-trained on 493 million satellite images, a domain strictly devoid of generative content.

\begin{table}[t]
\centering
\caption{\textbf{Counterfactual Analysis.} Comparison of DINOv3 trained on Web Data vs. Satellite Data.}
\label{tab:counterfactual}
\resizebox{0.75\linewidth}{!}{%
\begin{tabular}{l|c|ccc}
\toprule
\multirow{2}{*}{\textbf{Pre-training Data}} & \textbf{GenImage} & \multicolumn{3}{c}{\textbf{Chameleon}} \\
& Avg. & Real & Fake & Avg. \\
\midrule
\textbf{DINOv3-Web} & \textbf{0.965} & 0.933 & \textbf{0.895} & \textbf{0.914} \\
\textbf{DINOv3-Sat} & 0.706 & \textbf{0.948} & 0.121 & 0.535 \\
\bottomrule
\end{tabular}%
}
\vspace{-10pt}
\end{table}

Table~\ref{tab:counterfactual} delivers a decisive finding. While the web-trained baseline excels, DINOv3-Sat completely fails on fake images, despite performing well on real ones. This collapse proves that the model classifies unseen fakes as ``real" simply because it is not included in pretrained data. The conclusion is the forensic capability of SSL models is not inherent to the architecture or training strategy, but is entirely contingent on exposure to generative data during pre-training.


\section{Robustness and Limitations}

\subsection{Protocol I: Resilience to Common Perturbations}
\label{subsec:robust_perturbation}

To assess real-world reliability, we evaluate detector robustness against JPEG compression (Quality $\in \{95, \dots, 65\}$) and Gaussian Blur ($\sigma \in \{0.5, \dots, 2.0\}$). We benchmark modern VFM linear probes against legacy models and specialized detectors across both GenImage and the in-the-wild Chameleon datasets.

\begin{figure}[t]
    \centering
    \includegraphics[width=\linewidth]{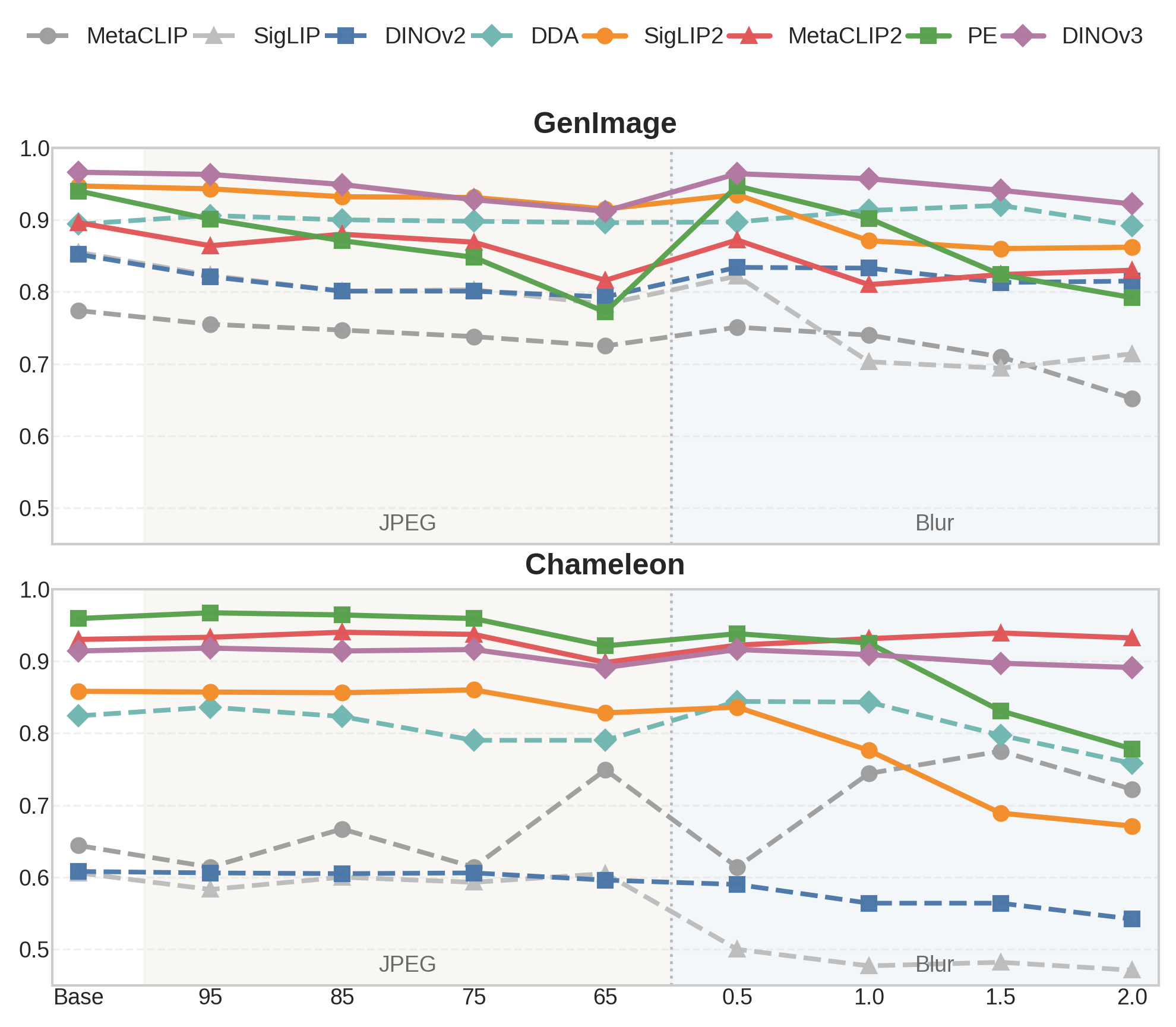} 
    \caption{\textbf{Robustness to Common Perturbations.} Accuracy trajectories under JPEG compression and Gaussian Blur.}
    \label{fig:robustness}
    \vspace{-5mm}
\end{figure}

As visualized in Figure~\ref{fig:robustness}, the trajectories reveal a stark stratification between legacy VFMs (dashed lines) and modern VFMs (solid lines). Older models not only establish lower baselines but also exhibit severe volatility under perturbations, particularly on Chameleon. This decisively proves that robustness is \textit{not} an inherent benefit of the linear probing architecture. Instead, modern VFMs exhibit superior resilience because their intrinsic forensic capabilities are derived from massive, inadvertent exposure to diverse synthetic content during web-scale pre-training. By internalizing the generative manifold directly from the messy, unconstrained internet, these models learn robust, high-level features rather than brittle, lab-generated artifacts. While \textbf{PE} still relies somewhat on high-frequency traces susceptible to low-pass filtering (dropping to 77.8\% at $\sigma=2.0$), \textbf{DINOv3} and \textbf{MetaCLIP2} capture fundamental structural anomalies that inherently resist smoothing. This profound resilience to blur and compression allows our image-trained linear probes (e.g., DINOv3) to achieve generalized SOTA performance on video benchmarks like VidProM and GenVideo via simple frame-level aggregation, decisively outperforming bespoke AI video detectors.

\subsection{Protocol II: Real-World Transmission and Recapture}
\label{subsec:robust_realworld}

To evaluate deployment reliability, we further assess performance under severe image degradation scenarios, including optical recapture from screens and heavy compression introduced by social media transmission protocols. To evaluate this, we utilize the RRDataset~\cite{li2025bridging}, measuring performance across three settings: \textbf{Original} (Digital baseline), \textbf{Redigital} (Screen or print recapture), and \textbf{Transfer} (Social media transmission).

\begin{table}[t]
\centering
\caption{\textbf{Robustness Evaluation on RRDataset.} We report the accuracy on Real and AI classes separately.}
\label{tab:rrdataset}
\resizebox{\linewidth}{!}{%
\begin{tabular}{l|cc|cc|cc}
\toprule
\multirow{2}{*}{\textbf{Detector}} & \multicolumn{2}{c|}{\textbf{Original (Base)}} & \multicolumn{2}{c|}{\textbf{Redigital (Recapture)}} & \multicolumn{2}{c}{\textbf{Transfer (Social App)}} \\
\cmidrule(lr){2-3} \cmidrule(lr){4-5} \cmidrule(lr){6-7}
& Real & AI & Real & AI & Real & AI \\
\midrule
CNNSpot & 0.977 & 0.189 & \textbf{0.986} & 0.068 & 0.993 & 0.065 \\
NPR & 0.957 & 0.749 & 0.955 & 0.083 & \textbf{0.998} & 0.002 \\
SAFE & 0.874 & 0.792 & 0.928 & 0.066 & \textbf{0.997} & 0.009 \\
DDA & 0.942 & 0.895 & 0.961 & 0.298 & 0.933 & 0.532 \\
\midrule
SigLIP2-Linear & 0.8714 & 0.931 & 0.885 & 0.559 & 0.490 & 0.704 \\
MetaCLIP2-Linear & 0.784 & \textbf{0.939} & 0.791 & \textbf{0.719} & 0.930 & \textbf{0.713} \\
PE-CLIP-Linear & 0.925 & \textbf{0.949} & 0.913 & 0.548 & \textbf{0.989} & 0.685 \\
DINOv3-Linear & \textbf{0.951} & 0.930 & \textbf{0.964} & 0.647 & 0.980 & \textbf{0.712} \\
\bottomrule
\end{tabular}%
}
\vspace{-10pt}
\end{table}

As detailed in Table~\ref{tab:rrdataset}, specialized detectors suffer a total collapse under transmission and recapture. Methods like SAFE and NPR degrade to near-zero sensitivity on fake images ($<1\%$). Even the robust DDA drops to 29.8\% on recaptured data.
In sharp contrast, modern VFMs maintain robust detection capabilities. \textbf{MetaCLIP2-Linear} leads with $\sim$72\% accuracy across both scenarios. Consistent with the blur experiments (Sec.~\ref{subsec:robust_perturbation}), \textbf{PE-CLIP} suffers a notable drop (to 54.8\%) on recaptured data, confirming its reliance on fine-grained details prone to erasure by low-pass filtering. Conversely, \textbf{DINOv3} and \textbf{MetaCLIP2} exhibit superior resilience, suggesting that their learned structural anomalies and semantic concepts persist even through the analog-to-digital bottleneck.


\subsection{Protocol III: Robustness to Reconstruction and Editing}
\label{subsec:robust_edit}

While modern VFMs excel at detecting fully generated images, forensic reliability requires handling subtler manipulations: \textbf{(1) VAE Reconstruction}, where a real image is encoded and decoded by a diffusion model's VAE without semantic modification (simulating deepfake pre-processing); and \textbf{(2) Local Editing}, where only specific regions are inpainted. We evaluate on the \textbf{DDA-COCO}~\cite{chen2025dual} (VAE-reconstructed real images) and \textbf{BR-Gen} ~\cite{cai2025zooming}(Diffusion-based local editing) datasets.

\begin{table}[t]
\centering
\caption{\textbf{Limitations under Reconstruction and Editing.} Detection accuracy on \textbf{DDA-COCO} (VAE-based reconstruction) and \textbf{BR-Gen} (Diffusion-based local editing).}
\label{tab:recon_edit}
\resizebox{0.85\linewidth}{!}{%
\begin{tabular}{l|ccc|ccc}
\toprule
\multirow{2}{*}{\textbf{Method}} & \multicolumn{3}{c|}{\textbf{DDA-COCO}} & \multicolumn{3}{c}{\textbf{BR-Gen}} \\
\cmidrule(lr){2-4} \cmidrule(lr){5-7}
& SDXL & SD2 & SD3.5L & Brush & Power & SDXL \\
\midrule
CNNSpot & 0.014 & 0.017 & 0.004 & 0.121 & 0.091 & 0.073 \\
DDA & \textbf{0.949} & \textbf{0.997} & \textbf{0.682} & 0.648 & 0.589 & 0.465 \\
OMAT & 0.211 & 0.304 & 0.135 & 0.697 & 0.715 & 0.681 \\
Effort & 0.511 & 0.549 & 0.682 & \textbf{0.801} & \textbf{0.793} & \textbf{0.767} \\
\midrule
SigLIP2-Linear & 0.071 & 0.079 & 0.017 & 0.597 & 0.623 & 0.476 \\
MetaCLIP2-Linear & 0.057 & 0.074 & 0.037 & 0.544 & 0.575 & 0.500 \\
PE-CLIP-Linear & 0.066 & 0.170 & 0.024 & 0.564 & 0.581 & 0.528 \\
DINOv3-Linear & 0.030 & 0.079 & 0.004 & 0.592 & 0.613 & 0.450 \\
\bottomrule
\end{tabular}%
}
\vspace{-10pt}
\end{table}

Table~\ref{tab:recon_edit} (Left) exposes a critical limitation: modern VFMs are essentially blind to pure VAE reconstruction artifacts. Detection rates plummet to negligible levels, indicating that these models do not perceive the low-level noise footprint of the VAE decoder as an anomaly. Conversely, \textbf{DDA}, which explicitly aligns with VAE reconstruction patterns, maintains robust performance. As shown in Table~\ref{tab:recon_edit} (Right), VFMs struggle to generalize to localized manipulations (BR-Gen), with performance hovering around \textbf{50\%--60\%}. We attribute this to the \textbf{global pooling mechanism} inherent to our linear probing approach: the dominant feature signal from the unaltered ``real'' regions likely suppresses the subtle forensic traces within the edited mask. In contrast, methods like \textbf{Effort}, designed to amplify anomaly feature, achieve higher accuracy.


\subsection{Protocol IV: The Bottleneck of Specialized Architectures}
\label{subsec:ablation_architectures}

A natural question arises from our primary finding: if modern Vision Foundation Models (VFMs) possess such powerful representations, could their performance be further amplified by attaching State-Of-The-Art specialized forensic architectures? To investigate this, we upgraded recent expert models—Effort \cite{yan2024orthogonal}, AIDE \cite{yan2024sanity}, and DDA \cite{chen2025dual}—by swapping their original legacy backbones with our top-performing modern VFMs (MetaCLIP2, PE, and DINOv3).

\begin{table}[ht]
\centering
\caption{\textbf{Impact of Upgrading Specialized Architectures.} While modern backbones improve the performance of specialized methods, they still severely underperform the simple frozen linear probe. Original baselines are in \textit{italics}, best results in \textbf{bold}.}
\label{tab:arch_upgrade}
\resizebox{\linewidth}{!}{
\begin{tabular}{l|ccc}
\toprule
\textbf{Method \& Backbone} & \textbf{GenImage} & \textbf{Chameleon} & \textbf{AIGIHolmes} \\
\midrule
\textit{Effort (Original: CLIP-L)} & \textit{0.807} & \textit{0.588} & \textit{0.722} \\
Effort + MetaCLIP2 & 0.748 & 0.720 & 0.799 \\
Effort + PE & 0.856 & 0.761 & 0.924 \\
\midrule
\textit{AIDE (Original: ResNet/ConvNeXt)} & \textit{0.768} & \textit{0.574} & \textit{0.970} \\
AIDE + MetaCLIP2 & 0.834 & 0.709 & 0.922 \\
AIDE + PE & 0.883 & 0.914 & 0.947 \\
\midrule
\textit{DDA (Original: DINOv2)} & \textit{0.890} & \textit{0.824} & \textit{0.963} \\
DDA + MetaCLIP2 & 0.557 & 0.563 & 0.596 \\
DDA + PE & 0.534 & 0.559 & 0.659 \\
DDA + DINOv3 & 0.572 & 0.751 & 0.713 \\
\midrule
\textbf{MetaCLIP2-Linear} & \textbf{0.892} & \textbf{0.930} & \textbf{0.942} \\
\textbf{PE-CLIP-Linear} & \textbf{0.938} & \textbf{0.959} & \textbf{0.978} \\
\textbf{DINOv3-Linear} & \textbf{0.964} & \textbf{0.914} & \textbf{0.972} \\
\bottomrule
\end{tabular}
}
\end{table}

The results in Table~\ref{tab:arch_upgrade} expose the \textbf{Bottleneck of Inductive Bias}. While upgrading expert models (e.g., AIDE, Effort) with modern VFMs improves their baseline performance, they still strictly underperform our minimalist linear probe (e.g., AIDE with PE achieves 91.4\% on Chameleon, falling short of PE-Linear's 95.9\%). Worse still, rigidly specialized architectures like DDA suffer catastrophic degradation ($\sim$50-60\%) when forced onto new generic feature spaces. This reveals that complex inductive biases—such as explicit frequency filtering or strict VAE alignment—actually act as information bottlenecks, inadvertently constraining the raw, universal discriminative power naturally emergent in modern representations.


\subsection{Protocol V: The Pitfall of Parameter-Efficient Fine-Tuning}
\label{subsec:ablation_lora}

Another prevalent paradigm for adapting foundation models is Parameter-Efficient Fine-Tuning (PEFT). If a simple linear layer suffices, would unfreezing the backbone via Low-Rank Adaptation (LoRA) yield even better task-specific performance? To test this, we applied LoRA with rank $r \in \{4, 8\}$ to modern VFMs, fine-tuning them on the GenImage (SD v1.4) training set.

\begin{table}[ht]
\centering
\caption{\textbf{LoRA Fine-Tuning vs. Frozen Linear Probe.} Unfreezing the backbone via LoRA significantly degrades generalization in the wild. Best results in \textbf{bold}.}
\label{tab:lora_ablation}
\resizebox{\linewidth}{!}{
\begin{tabular}{ll|ccc}
\toprule
\textbf{Backbone} & \textbf{Strategy} & \textbf{GenImage} & \textbf{Chameleon} & \textbf{AIGIHolmes} \\
\midrule
\multirow{3}{*}{\textbf{MetaCLIP2}} 
& \textbf{Linear Probe (Frozen)} & \textbf{0.892} & \textbf{0.930} & \textbf{0.942} \\
& LoRA (r=4) & 0.734 & 0.817 & 0.823 \\
& LoRA (r=8) & 0.780 & 0.880 & 0.896 \\
\midrule
\multirow{3}{*}{\textbf{PE}} 
& \textbf{Linear Probe (Frozen)} & \textbf{0.938} & \textbf{0.959} & \textbf{0.978} \\
& LoRA (r=4) & 0.810 & 0.719 & 0.879 \\
& LoRA (r=8) & 0.761 & 0.635 & 0.891 \\
\midrule
\multirow{3}{*}{\textbf{DINOv3}} 
& \textbf{Linear Probe (Frozen)} & \textbf{0.964} & \textbf{0.914} & 0.972 \\
& LoRA (r=4) & 0.954 & 0.803 & \textbf{0.977} \\
& LoRA (r=8) & 0.928 & 0.718 & 0.945 \\
\bottomrule
\end{tabular}
}
\vspace{-10pt}
\end{table}

Table~\ref{tab:lora_ablation} highlights the \textbf{Risk of Modifying Internal Knowledge}. Contrary to the intuition that fine-tuning improves task adaptation, applying LoRA actively dismantles generalizability. For instance, LoRA fine-tuning on PE (r=8) causes its in-the-wild detection accuracy (Chameleon) to plummet from 95.9\% to a mere 63.5\%. We attribute this to \textit{manifold distortion} and \textit{catastrophic forgetting}. By unfreezing the backbone to optimize for a single generator (SD v1.4), the model rapidly overfits to a narrow, transient distribution of local artifacts, irrevocably overwriting the broad "world knowledge" of synthetic anomalies it implicitly derived from massive pre-training data.

\section{Conclusion}
\label{sec:conclusion}
In this work, we revisit AIGI detection from the perspective of pretrained visual representations. We show that frozen features from modern Vision Foundation Models, combined with a lightweight classifier, form a remarkably strong baseline for generalizable AIGI detection. Across standard benchmarks, in-the-wild datasets, and recent unseen generators, this simple setup consistently matches or outperforms recent specialized detectors. Our analyses further suggest that this capability is closely related to exposure to synthetic web content during pre-training, rather than primarily to forensic-specific architectural design. In VLMs, this appears as semantic alignment with forgery-related concepts, while in SSL models it appears as implicit discrimination of generative distributions. Although fully controlled pre-training ablations are beyond the scope of this work, our evidence consistently supports this interpretation.

At the same time, modern VFMs remain weak on pure VAE reconstruction, localized editing, and severe transmission or recapture. We therefore view frozen modern VFM representations not as a complete solution to multimedia forensics, but as a strong foundation for robust global AIGI detection. More broadly, our findings suggest that future progress may depend less on increasingly specialized detector design, and more on effectively leveraging the evolving representations learned by foundation models.

\bibliographystyle{ACM-Reference-Format}
\bibliography{icml}
\end{document}